# Can AI Tools Transform Low-Demand Math Tasks? An Evaluation of Task Modification Capabilities

Danielle S. Fox[1], Brenda L. Robles[2], Elizabeth DiPietro Brovey[2], Christian D. Schunn[1,2]

[1] Learning Research and Development Center, University of Pittsburgh

[2]Institute for Learning, University of Pittsburgh

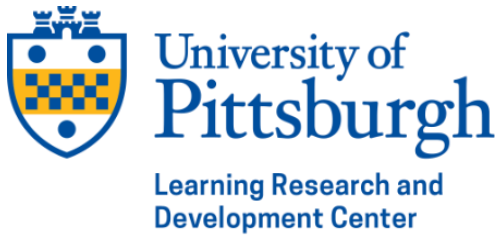
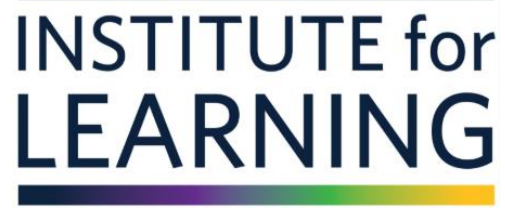

# Abstract

While recent research has explored AI tools' ability to classify the quality of mathematical tasks (Fox et al., 2026), little is known about their capacity to increase the quality of existing tasks. This study investigated whether AI tools could successfully upgrade low-cognitive-demand mathematics tasks. Eleven tools were tested, including six broadly available, general-purpose AI tools (e.g., ChatGPT and Claude) and five tools specialized for mathematics teachers (e.g., Khanmigo, coteach.ai). Using the Task Analysis Guide framework (Stein & Smith, 1998), we prompted AI tools to modify two different types of low-demand mathematical tasks. The prompting strategy aimed to represent likely approaches taken by knowledgeable teachers, rather than extensive optimization to find a more effective prompt (i.e., an optimistic typical outcome). On average, AI tools were only moderately successful: tasks were accurately upgraded only 64% of the time, with different AI tool performance ranging from quite weak (33%) to broadly successful (88%). Specialized tools were only moderately more successful than general-purpose tools. Failure modes included both "undershooting" (maintaining low cognitive demand) and "overshooting" (elevating tasks to an overly ambitious target category that likely would be rejected by teachers). Interestingly, there was a small negative correlation ($r$ = - .35) between whether a given AI tool was able to correctly classify the cognitive demand of tasks and whether the AI was able to upgrade tasks, showing that the ability to modify tasks (i.e., a generative task) represents a distinct capability from the ability to classify them (i.e., judgement using a rubric). These findings have important implications for understanding AI's potential role in curriculum adaptation and highlight the need for specialized approaches to support teachers in modifying instructional materials.

# 1. Introduction

## 1.1 The Challenge of Curriculum Adaptation

Teachers rarely serve their students well by implementing curricula exactly as written. Contextual factors like time constraints, student readiness, available resources, and pacing demands require constant adaptation of instructional materials (Remillard, 2005; Zhou & Lo, 2025). In mathematics education, one critical adaptation involves modifying tasks to achieve appropriate levels of cognitive demand (Stein & Smith, 1996). Many curricula include too many low-demand tasks that focus on memorizing algorithms or procedural execution (Boston & Smith, 2009; Stein et al., 1996), and teachers need to modify them to promote deeper conceptual understanding and mathematical reasoning. However, such modifications require advanced pedagogical content knowledge and time (Agyapong et al., 2022; Creagh et al., 2025; Monarrez & Tchoshanov, 2022; Wang, 2024). Teachers must understand not only the mathematical content but also how different task features influence student thinking. In particular, they must identify which elements of a task reduce cognitive demand (e.g., explicitly stating an algorithm to use to solve the problem; Tekkumru-Kisa & Stein, 2020) and determine how to modify those elements while maintaining mathematical integrity and coherence with learning goals (Boston & Smith, 2009; Tekkumru-Kisa & Stein, 2015). This is time-intensive work that competes with teachers' many other responsibilities.

## 1.2 AI as a Potential Solution

Artificial intelligence tools have shown promise in various educational applications, from tutoring systems to automated grading. Recent advances in large language models (LLMs) raise the question of whether AI might also support teachers in curriculum adaptation tasks, including modifying the cognitive demand of mathematical tasks. General-purpose tools are now widely available to teachers for personal and career use, including instructional planning. In addition, many specialized tools have been created specifically to support mathematics teachers. If AI tools can reliably upgrade low-demand tasks to promote deeper thinking, they could reduce teacher workload while improving instructional quality. However, we currently lack empirical evidence about AI tools' capacity for this type of pedagogical transformation. Prior research has demonstrated that AI can generate educational content, but generating new content differs fundamentally from strategically modifying existing materials to achieve specific pedagogical goals. Further, AI may often generate incomplete solutions or ones that are poorly customized to a specific context (Cao et al., 2026).

## 1.3 Research Gaps

Our understanding of AI capabilities for task modification faces several limitations:

1. **Limited empirical evidence:** Most studies of AI in education focus on student-facing intelligent tutoring (Hwang & Tu, 2021), or more generally, content generation, rather than strategic modification of existing materials.

2. **Unclear relationship to classification ability:** AIs appear to vary substantially in their ability to classify the cognitive demand of given tasks (Fox et al., 2026). In humans, the ability to classify correctly is considered foundational to generating or modifying correctly. However, it is not clear that LLMs will also be limited in this way, given their ability to retrieve prior examples of cognitively demanding tasks.

3. **Insufficient analysis of failure modes:** When AI tools fail to modify tasks successfully, we do not know whether certain error types are more common (e.g., undershooting the target demand level, overshooting the target demand level, or whether they make other types of errors, such as failing to address other classroom needs).

4. **Lack of tool comparison:** We do not know the extent of performance variation among currently available AI tools, particularly whether education-specific AI tools will outperform general-purpose AI tools for task modification.

## 1.4 Theoretical Framework

This study employs the Task Analysis Guide (TAG) developed by Stein and Smith (1998), a research-backed framework that continues to be used for mathematics teacher education and professional development (see Appendix A). The TAG categorizes mathematical tasks into four levels of cognitive demand:

**Low Cognitive Demand:**
1. **Memorization**: Tasks that require reproduction of previously learned facts, rules, formulas, or definitions without connections to concepts or meaning (e.g., *What are the decimal and percent equivalents for the fractions ½ & ¼?*)

2. **Procedures without Connections**: Tasks that focus on producing correct answers using procedures without developing conceptual understanding (e.g., *Convert the fraction 3/8 to a decimal and a percent. Show your work.*)

**High Cognitive Demand:**
3. **Procedures with Connections**: Tasks that focus students' attention on the use of procedures for the purpose of developing a deeper understanding of mathematical concepts and ideas (e.g., *Using a 10 x 10 grid, identify the decimal and percent equivalents of 3/8.*)

4. **Doing Mathematics**: Tasks that require complex, non-algorithmic thinking and demand self-monitoring of one's own cognitive processes (e.g., *Shade 6 of the small squares in the rectangle below.*

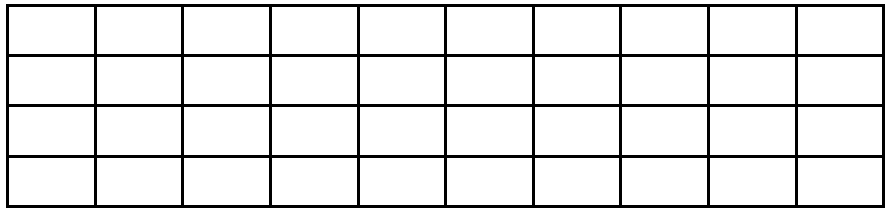

*Using the diagram, explain how to determine each of the following:*
- *the percent of the area that is shaded*
- *the decimal part of the area that is shaded*
- *the fractional part of the area that is shaded)*

In the present study, we focus on modifications targeting the *Procedures with Connections* category because, while *Doing Mathematics* tasks represent the highest level of cognitive demand, they are also difficult to implement effectively given the previously mentioned contextual limitations (e.g., time; Parrish & Byrd, 2022; Stein et al., 1996). Whereas the *Procedures with Connections* represents a high cognitive-demand task type that teachers are more likely to think will meet the needs of their classroom and will have time to implement effectively. Tasks at this level maintain procedural scaffolding while requiring students to make meaningful connections to underlying mathematical concepts.

## 1.5 Research Questions

This study addresses the following questions:

1. Can current general and specialized education AI tools successfully modify low-cognitive-demand mathematics tasks to meet the criteria for *Procedures with Connections*?
2. How does modification success vary across different AI tools?

3. How does modification success vary across different types of low-demand tasks (*Memorization* vs. *Procedures without Connections*)?

4. What is the relation between an AI tool's ability to accurately classify cognitive demand and its ability to accurately modify tasks?

5. What patterns of failure (undershooting vs. overshooting) characterize unsuccessful modification attempts?

# 2. Methods

## 2.1 Building on Prior Classification Work

### 2.1.1 Methods

This study builds directly on our prior investigation of AI tools' ability to classify cognitive demand (Fox, et al., 2026). That study tested the ability of 11 AI tools to classify mathematics tasks using the Task Analysis Guide. The current study uses the same 11 tools and focuses on six tasks classified by human experts as low-cognitive demand (for complete task information, see Appendix B):

Tasks were renamed to reflect the underlying category.

**Memorization:** Tasks M1, M2, M3

Example: Task M1

*What are the decimal and percent equivalents for the fractions ½ and ¼?*

Classification Rationale: This task, as written, is asking students to produce previously learned facts from memory and has no connection to the underlying mathematical concepts.

**Procedures without Connections:** Tasks PwoC1, PwoC2, PwoC3

Example: Task PwoC1

*Convert the fraction 3/8 to a decimal and a percent. Show your work.*

Classification Rationale: This task is algorithmic in nature and requires limited cognitive demand for completion. While a procedure is required to complete the task rather than just reproducing previously learned facts, there is no connection to the mathematical concepts that underlie the steps. There is no ambiguity about what needs to be done, and no explanation is required; only showing the steps taken to complete it.

## 2.2 AI Tool Selection

### 2.2.1 Methods

We tested the same 11 AI tools from our classification study (Fox et al., 2026):

**General-Purpose AI Tools (n = 6):**
- ChatGPT GPT 5
- Claude Sonnet 4.5
- DeepSeek -V3.2-Exp
- Gemini Free 2.5 Flash
- Grok Free 4.1
- Perplexity

**Education-Specific AI Tools (n = 5):**
- Brisk: Beta
- Coteach AI: Free
- Khanmigo: Free
- Magic School: Free
- School.AI: Free

Note: Brisk and Khanmigo had technical limitations that prevented them from completing all modification tasks, resulting in incomplete data for these tools.

## 2.3 Prompting Procedure

For each of the six low-demand tasks, we provided each AI tool with:

1. The Task Analysis Guide (TAG.docx)
2. The original task (Task X.docx)
3. A standardized modification prompt

**Standardized Prompt:**

*"Based on the TAG.docx I uploaded, Task [X].docx is considered a lower cognitive demand task (Memorization/Procedures without Connections). Modify this task so that it meets the Procedures with Connections classification. Use the TAG.docx as the framework to modify the task. Give detailed reasoning and output the detailed analysis with dimensions in a table."*

This simple prompting approach was meant to capture the output quality a teacher would produce with knowledge of the task analysis guide and some basics of prompt engineering. That is, we did not employ extensive prompt engineering techniques that would be likely beyond the resources of most teachers. This performance baseline could, however, inform future work aimed at new tools that embed more effective prompts to further improve outcomes.

## 2.4 Human Expert Evaluation

Each AI-modified task was evaluated by human experts trained in the use of the Task Analysis Guide (see Appendix C for the complete coding protocol). Experts assessed whether each modification should be classified as an:

- **Undershoot:** The modified task remained in the low-cognitive-demand categories (*Memorization* or *Procedures without Connections*), failing to achieve the target level.
- **Success:** The modified task met the criteria for *Procedures with Connections*, achieving the target cognitive demand level.
- **Overshoot:** The modified task exceeded the target, meeting the criteria for *Doing Mathematics* (the highest cognitive demand category).

While overshooting might seem like a positive outcome, it represents a failure to meet the specified objective. Teachers targeting *Procedures with Connections* may have good pedagogical reasons for that choice as these tasks provide more scaffolding than *Doing Mathematics* tasks while still promoting conceptual understanding. Tasks in the *Procedures with Connections* category may also be used if there is insufficient time to work through a higher-demand task. Thus, an overshot task may be inappropriate for the intended instructional context and time.

## 2.5 Data Analysis

We calculated success rates at multiple levels:

- Overall success rate (i.e., percentage of modifications coded as meeting the *Procedures with Connections* level)
- Success rate by AI tool
- Success rate by task
- Success rate by original task type (*Memorization* vs. *Procedures without Connections*)
- Undershoot and Overshoot rates by AI tool

We also examined the correlation between each tool's classification accuracy (from our prior study) and modification success.

# 3 Results

Across all tools and tasks, AI tools successfully upgraded low-cognitive-demand tasks to meet the criteria for *Procedures with Connections* 64% of the time. This represents modest success; better than chance but far from reliable autonomous performance. Additionally, there was substantial variation across both tools and tasks.

## 3.1 Performance by AI Tool

Modification success varied considerably across AI tools, ranging from 33% to 83% (see Table 1). In general, accuracy was similar between education-specific tools ($M$ = 65%) and general AI tools ($M$ = 61%).

**Table 1.** *AI task modification success in decreasing order of success. Education-specific tools in bold.*

| AI Tool (base tool) | Upgrade Success Rate | Undershot Rate | Overshot Rate | Classification Accuracy |
|---|---|---|---|---|
| Gemini | 83% | 17% | 0% | 58% |
| **School.AI (Open AI ChatGPT)** | 83% | 0% | 17% | 58% |
| **Khanmigo (Microsoft)** | 75%* | 25% | 0% | 50% |
| **Brisk (Google Gemini)** | 67%* | 33% | 0% | 65% |
| DeepSeek | 67% | 33% | 0% | 83% |
| **Magic School (Multi-modal)** | 67% | 33% | 17% | 67% |
| Perplexity (Multi-modal) | 66% | 17% | 17% | 58% |
| Claude | 66% | 17% | 17% | 50% |
| ChatGPT | 50% | 0% | 50% | 50% |
| **Coteach (Anthropic Claude)** | 33% | 17% | 50% | 75% |
| Grok | 33% | 57% | 0% | 67% |

*Notes: * Brisk and Khanmigo could not complete all tasks due to technical limitations; their success rates are calculated from available cases.*

**Top Performers:** Gemini and School.AI achieved 83% success, making them the most reliable tools for this task.

**Frequent Overshooting:** ChatGPT and Coteach frequently overshot the target, creating tasks that exceeded the desired cognitive demand level (50% overshoot rate).

**Conservative Approach:** Several tools (Gemini, DeepSeek, Brisk, Khanmigo, Grok) never overshot, suggesting these tools may be more conservative in their modifications.

## 3.2 Performance by Task

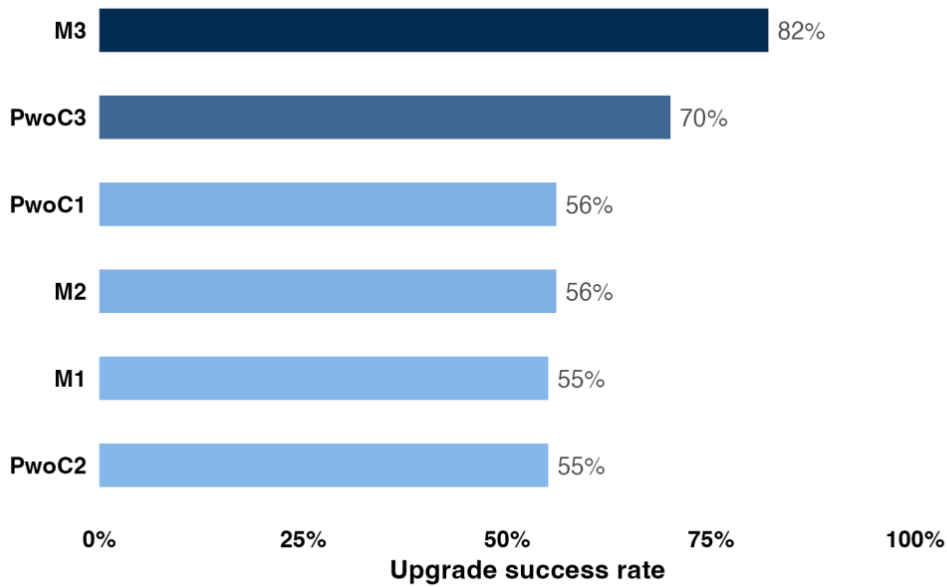

*Figure 1. Task upgrade success rate by task, in decreasing order.*

Success rates also varied across tasks (see Figure 1), with two tasks being upgraded more often than the other four. Interestingly, the original task category (*Memorization* vs. *Procedures without Connections*) did not systematically predict the success of modification. The mean success rate was 64% both task types (For complete modification data, see Appendix D).

## 3.3 Relation between Classification success and Modification success

Surprisingly, the correlation between classification accuracy and modification success was negative (*r* = -.35). These results should be interpreted with caution, given the small number of cases being examined. However, these results show that classification skills are not the main source of failure to upgrade tasks. Classification requires using a rubric to recognize and parse features of existing tasks, while modification requires generating new content that achieves specific pedagogical goals.

**Examples**:
- **DeepSeek:** Highest classification accuracy (83%) but moderate modification success (67%)
- **Coteach:** High classification accuracy (75%) but lowest modification success (33%)
- **Gemini:** Moderate classification accuracy (58%) but highest modification success (83%)
- **School.AI:** Moderate classification accuracy (58%) but highest modification success (83%)

## 3.4 Patterns of Failure – Undershooting vs. Overshooting

When AI tools failed to achieve the target (*Procedures with Connections*), they failed in different ways:

**Undershooting (maintaining low cognitive demand):**
- More common error

- Occurred when tools made only minor modifications that did not fundamentally change cognitive demands (e.g., adding context without changing the nature of student thinking required, or reformulating the same procedural task in slightly different words)

**Example**: Magic School's modification of Task M1 (originally *Memorization*)

**Original Task M1**:

*What are the decimal and percent equivalents for the fractions 1/2 and 1/4?*

**Magic School's modification of Task M1:**

*Given the fractions 3/4 and 2/5:*

*1. Use long division to convert each fraction to its decimal equivalent. Show your work and explain how the division process relates to the meaning of the fraction.*

*2. Then, convert each decimal to a percent by explaining the connection between decimals and percentages.*

*3. Represent each fraction visually (e.g., with a pie chart or number line) and explain how this representation supports your conversions.*

Magic School's modification of Task M1 failed to meet the criteria for *Procedures with Connections* and instead created a task that aligns with the criteria for *Procedures without Connections*. In particular, the modified task is algorithmic in that it explicitly directs students to use long division (i.e., a single, prescribed algorithm) to convert fractions to decimals in step one, failing to provide an opportunity for students to discover the connection between percentages and decimals. In addition, the tool changed the fractions in the task from 1/2 and 1/4 to 3/4 and 2/5, which represents a failure to adapt the task with fidelity to the original.

There is a clear attempt to increase the cognitive demand by asking students to explain their work and make connections between procedures and representations; however, these additions are considered "window dressing" because converting fractions to decimals via long division and then to a percent does not actually get at the meaning of the fractions or conceptual connections to the underlying mathematics. Overall, this modification is more representative of a task at the *Procedures without Connections* level, as there is little ambiguity about how to solve it.

**Overshooting (exceeding the target to *Doing Mathematics*):**
- Varied dramatically across tools (0% to 50%)
- ChatGPT overshot 50% of the time, making tasks too complex for the target level
- Coteach also showed high overshooting (50%)
- Six tools never overshot (Gemini, DeepSeek, Brisk, Khanmigo, Grok, Magic School), but it is important to note that Brisk and Khanmigo were unable to complete all of the modifications.

**Example**: Coteach's modification of Task M2 (Originally *Memorization*)

**Original Task M2**:

*True or False, 4/100 < 4/96*

**Coteach's modification of Task M2:**

*The School Garden Project*

*Two classes are working on school garden projects:*

*Class A: Has completed 4 out of 100 planned garden squares*

*Class B: Has completed 4 out of 96 planned garden squares*

*Part 1: Initial Comparison*

> 1. Without calculating decimal values, predict which class has completed a greater fraction of their garden project. Explain your reasoning using what you know about fractions.
> 2. Create visual models (such as grids or bar diagrams) to represent both fractions. What do you notice when you compare them visually?
>
> *Part 2: Mathematical Analysis*
>
> 3. Use at least two different mathematical methods to determine whether
>
> 4/100 < 4/96. For each method, explain: Why the method works mathematically, how it connects to the meaning of fractions, what it reveals about the relationship between numerators and denominators
>
> 4. Convert both fractions to equivalent fractions with the same denominator.
>
> *Explain how this method helps you understand why one fraction is larger than the other.*
>
> *Part 3: Pattern Recognition and Generalization*
>
> 5. Consider the general case: If you have two fractions with the same numerator (like a/b and a/c), how can you determine which is larger without doing any calculations? Test your rule with other examples.
> 6. Explain why your rule works by connecting it to the real-world garden context. What does it mean for the denominator to be larger when the numerator stays the same?
> 7. Create your own comparison problem similar to this one and explain how someone could solve it using the insights you've developed.

This modification is a clear overshoot by Coteach as it increased the cognitive demand level beyond the *Procedures with Connections* level into the *Doing Mathematics* classification. It requires students to access their prior learning about fractions, use non-algebraic thinking and self-regulation while making a prediction, and then verify their thinking by creating visual representations and solving the problem using two methods. Students must also create a generalizable rule and generate their own problem; extending their understanding and making connections between the mathematics and a real-world problem. Together, the steps require students to make connections to the underlying mathematical concepts of fractions and to reason about how the denominator affects the size of the parts.

Additional examples of AI-modified tasks can be found in Appendix E.

## 3.5. Comparing Modification Patterns Across Tools

The distinct patterns by tool suggest different approaches. Some tools (ChatGPT, Coteach) appear to take an aggressive approach, making substantial modifications that risk overshooting. Others (Gemini, School.AI) appear to take a more conservative approach, making measured changes that avoid overshooting but occasionally undershooting.

*High Success:*
- Gemini, School.AI: 83% success, 0% overshoot, 17% undershoot
- These tools demonstrate the most reliable modification capability

*Moderate Success, Low Overshoot (Conservative Pattern):*
- DeepSeek, Claude, Perplexity, Magic School: 67% success, 0-17% overshoot, 17%-33% undershoot
- Reliable but less consistent than top performers

*Moderate Success, High Overshoot (Aggressive Pattern):*
- ChatGPT: 50% success, 50% overshoot
- Coteach: 33% success, 50% overshoot

- These tools made dramatic changes that often exceeded the target

*Low Success (Undershoot Pattern):*
- Grok: 33% success, 67% undershoot
- This tool tends to make insufficient modifications

# 4. Discussion

## 4.1 Interpretation of Findings

The 64% overall success rate presents a nuanced picture of AI capabilities for task modification. This performance level suggests that AI tools can meaningfully engage with the challenge of elevating cognitive demand, but they are not yet reliable enough for autonomous deployment without human oversight.

The substantial variation across tools (33% – 83%) indicates that tool selection matters. Teachers or researchers seeking to use AI for task modification should carefully evaluate different tools rather than assuming equivalent performance across platforms. It was also clear that using an education-specific tool was not more helpful for this task.

## 4.2 The Potential Classification-Modification Trade-Off

Perhaps the most theoretically interesting finding is the clear lack of positive correlation between classification accuracy and modification success. This counterintuitive pattern challenges the assumption that tools good at recognizing cognitive demand would also be good at manipulating it. Given that the present study was based on a small number of cases, this trend warrants replication with a larger sample. Several explanations might account for this pattern:

**Different Processes:** Classification may rely primarily on pattern matching and feature recognition, while modification requires generative capabilities and creative problem-solving. These may draw on different aspects of AI architecture.

**Risk-taking vs. Conservatism**: Tools optimized for classification accuracy may be more conservative in their modifications, avoiding substantial changes that could produce errors. In contrast, tools willing to make bolder modifications might achieve higher success when they hit the target, even if they're less accurate at classification.

**Training Data Differences**: Tools may have been trained on different types of educational content, with some seeing more examples of task modification and others seeing more examples of task analysis.

**Overfitting Classification**: Tools highly optimized for classification might overly rely on surface features that don't translate to effective modification strategies.

**Practical Implications:** Educators cannot assume that a tool's classification performance predicts its modification performance. These capabilities must be evaluated independently.

## 4.3 Understanding Overshooting vs. Undershooting

The dramatic differences in overshooting rates across tools (0% to 50%) reveal distinct modification strategies:

Conservative tools (Gemini, School.AI, DeepSeek, Grok, Brisk, Khanmigo) never or rarely overshoot. These tools appear to make measured modifications, potentially focusing on minimal changes necessary to achieve the target level. While this reduces the risk of exceeding the target, it may also increase the risk of undershooting.

Aggressive tools (ChatGPT, Coteach) frequently overshoot, suggesting they make substantial modifications that often go beyond what was requested. This might reflect training that emphasizes maximizing student thinking or difficulty, or it might reflect difficulty calibrating the extent of modifications needed.

From a practical standpoint, the error type matters. Undershooting means the modified task still fails to achieve the desired cognitive demand, wasting the modification effort. Overshooting means the task may be too demanding for the intended context, potentially frustrating students or requiring more time than available. Neither is ideal, but depending on context, teachers might prefer one error type over the other.

## 4.4 Task Characteristics Matter

The variation in success rates across tasks (55% to 82%) indicates that certain tasks are more amenable to cognitive demand elevation than others. The 82% success rate for Task M3 suggests it had features that made the pathway to *Procedures with Connections* more apparent or achievable.

Examining Task M3 more closely might reveal:

- Clear opportunities for adding conceptual connections
- Mathematical content that naturally lends itself to deeper investigation
- Structure that allows modification without complete redesign

In contrast, tasks with ~55% success rates may have been constrained by:

- Content that resists easy connection to deeper concepts
- Format or structure that would require complete reconstruction
- Ambiguity about what conceptual connections would be appropriate

Understanding what makes tasks easy or difficult to modify could inform both AI development and human curriculum design.

## 4.5 Practical Implications

These findings have several implications for educational practice:

**Tool Selection:** The 50-percentage-point spread between the worst and best performers (33% – 83%) underscores that choosing the right tool substantially impacts outcomes. Based on this study, Gemini and School.AI appear to be the most reliable for task modification.

**Human Oversight is Essential:** Even the best tools succeeded only 83% of the time. Teachers should review and refine AI-modified tasks rather than using them directly.

**Error Preferences:** Teachers might select tools based on their overshooting vs. undershooting patterns, depending on whether they prefer tasks that exceed vs. fall short of the target.

**Different Tools, Different Purposes:** The small, negative correlation between classification and modification success, though insignificant and to be interpreted with caution, suggests that optimal workflows might use different tools for different purposes: one tool for classification, and another for modification. This trend parallels findings from AI leaderboards (Arena, 2026), which show that different tools have strengths in some domains (e.g., text, coding, vision, text-to-image) but not across all domains.

Teachers might use classification tools to identify low-demand tasks needing modification, then use modification tools to elevate them, with final human review.

## 4.6 Future Directions

The present study highlights several opportunities for future research.

**Task Sample and Diversity:** The current sample of six tasks, while purposefully selected to represent different types of low-demand tasks, is limited in scope. Future work should test a broader, more diverse set of tasks spanning more topics, grade levels, and task types.

**Beyond Procedures with Connections:** This study only examined modifications toward *Procedures with Connections*, leaving open questions about whether AI tools can successfully modify or create tasks to align with other levels of cognitive demand.

**Prompt Engineering and Revision:** The baseline prompting approach we employed in the present study deliberately avoided optimization techniques. Ongoing work is investigating whether carefully engineered prompts and iterative refinement (allowing tools to make revisions based on feedback) can substantially improve performance (Fox et al., in prep).

**Ensemble Approaches:** Using multiple tools or combination approaches may also achieve higher success rates than any single tool alone.

**More Detailed Analyses of Modified Task Quality:** There are many dimensions that make tasks effective in instruction, and future analyses should examine dimensions beyond those examined here. For example, AI can provide implausible real-world cases, inaccurate diagrams, and errors in mathematical formulas (Cao et al., 2026).

**Inter-rater Reliability:** On the measurement side, future work should establish inter-rater reliability for modification evaluations to reduce uncertainty around borderline cases, and longitudinal studies should track whether these patterns remain stable as AI models continue to evolve.

**Complete Dataset:** Technical limitations prevented Brisk and Khanmigo from completing all tasks in the current study; ensuring complete data across all tools will be important for drawing fuller conclusions in future work.

# 5. Conclusion

This study provides the first systematic evaluation of AI tools' capacity to modify mathematics tasks to achieve higher cognitive demand. The 64% average success rate demonstrates that many current AI tools struggle with this pedagogical challenge, and the substantial variation across tools (33% – 83%) indicates that thoughtful tool selection is crucial.

The small, negative correlation between classification accuracy and modification success ($r$ = -.35) hints that these may represent distinct capabilities, but more investigation is needed to unpack and fully characterize these different skills. The divergent patterns of overshooting versus undershooting across tools further suggest distinct modification strategies, with some tools taking conservative approaches and others making more aggressive changes.

Gemini and School.AI emerged as the most reliable tools for this task, achieving 83% success with no overshooting. However, even these top performers fell short of the reliability levels needed for autonomous deployment. Current AI tools are best positioned as decision-support systems that augment teacher expertise rather than replacing teacher judgment.

As AI becomes increasingly integrated into educational contexts, understanding both capabilities and limitations for specific pedagogical tasks is essential. This research demonstrates that AI tools show promise for supporting curriculum adaptation while highlighting the continued indispensable role of teacher expertise in creating and refining instructional materials that meet students' learning needs.

# Appendix A: Mathematical Task Analysis Guide

| **Lower-Level Demands** | **Higher-Level Demands** |
|---|---|
| *Memorization Tasks* <br>• Involves either producing previously learned facts, rules, formulae, or definitions OR committing facts, rules, formulae, or definitions to memory. <br>• Cannot be solved using procedures because a procedure does not exist or because the time frame in which the task is being completed is too short to use a procedure. <br>• Are not ambiguous – such tasks involve exact reproduction of previously seen material and what is to be reproduced is clearly and directly stated. <br>• Have no connection to the concepts or meaning that underlie the facts, rules, formulae, or definitions being learned or reproduced. | *Procedures with Connections Tasks* <br>• Focus students' attention on the use of procedures for the purpose of developing deeper levels of understanding of mathematical concepts and ideas. <br>• Suggest pathways to follow (explicitly or implicitly) that are broad general procedures that have close connections to underlying conceptual ideas as opposed to narrow algorithms that are opaque with respect to underlying concepts. <br>• Usually are represented in multiple ways (e.g., visual diagrams, manipulatives, symbols, problem situations). Making connections among multiple representations helps to develop meaning. <br>• Require some degree of cognitive effort. Although general procedures may be followed, they cannot be followed mindlessly. Students need to engage with the conceptual ideas that underlie the procedures in order to successfully complete the task and develop understanding. |
| *Procedures without Connections Tasks* <br>• Are algorithmic. Use of the procedure is either specifically called for, or its use is evident based on prior instruction, experience, or placement of the task. <br>• Require limited cognitive demand for successful completion. There is little ambiguity about what needs to be done and how to do it. <br>• Have no connection to the concepts or meaning that underlie the procedure being used. <br>• Are focused on producing correct answers rather than developing mathematical understanding. <br>• Require no explanations, or explanations that focus solely on describing the procedure that was used. | *Doing Mathematics Tasks* <br>• Requires complex and non-algorithmic thinking (i.e., there is not a predictable, well-rehearsed approach or pathway explicitly suggested by the task, task instructions, or a worked-out example). <br>• Requires students to explore and to understand the nature of mathematical concepts, processes, or relationships. <br>• Demands self-monitoring or self-regulation of one's own cognitive processes. <br>• Requires students to access relevant knowledge and experiences and make appropriate use of them in working through the task. <br>• Requires students to analyze the task and actively examine task constraints that may limit possible solution strategies and solutions. <br>• Requires considerable cognitive effort and may involve some level of anxiety for the student due to the unpredictable nature of the solution process required. |

# Appendix B: Original Low-Demand Tasks (M1-3, PwoC1-3)

**Task M1:**

*What are the decimal and percent equivalents for the fractions ½ and ¼?*

**Task M2:**

*True or False? 4/100 < 4/96*

**Task M3:**

*In the following equations, name the slope and y-intercept.*

   A.   $y = 8x + 15$
   B.   $y = -2x + 7$
   C.   $y = 4x - 9$

**Task PwoC1:**

*Convert the fraction 3/8 to a decimal and a percent. Show your work.*

**Task PwoC2:**

*Use cross products to solve the proportion. Show your work.*
*4/96 = x/6000*

**Task PwoC3:**

*If $y = 8x + 15$, evaluate y when x =*

   A.   10
   B.   20
   C.   30

# Appendix C: Expert Coding Protocol

To evaluate each of the AI modified tasks, three experts trained in using the Task Analysis Guide independently reviewed one third of the modified tasks (22 tasks each), blind to one another's scoring, and systematically determined if each of the criteria for *Procedures with Connections* was met. The following steps were taken by each reviewer for each task:

1. Compare the original task to the modified task to ensure that the goal of the task (e.g., comparing fractions), numbers, and operations were consistent. If the AI tool changed any of the core components during modification (e.g., the numbers used in the problems), then it was marked as a failure.

2. Note specific lines of text within the adapted task (e.g., "Use long division to solve...") that related to the criteria for *Procedures with Connections* tasks (e.g., "Suggest pathways to follow (explicitly or implicitly) that are broad general procedures...").

3. Review the criteria for every cognitive demand level of the Task Analysis Guide. Note specific lines of the adapted task that connect to criteria across different levels of cognitive demand. For example, if the modified task is very cognitively demanding but does not provide an explicit or implicit solution pathway, then it aligns more with the first criterion of the *Doing Mathematics* classification:

    > Requires complex and non-algorithmic thinking (i.e., there is not a predictable, well-rehearsed approach or pathway explicitly suggested by the task, task instructions, or a worked-out example).

4. Final judgment is made to decide if the adapted task meets, exceeds, or undershoots the goal of *Procedures with Connections*.

After each reviewer had independently evaluated all their assigned modifications, the reviewers met to compare notes and evaluations of all 66 tasks and made a final determination together. If disagreement occurred among reviewers, the reviewer with the most experience made the final determination.

# Appendix D: Complete Modification Evaluation Data

|  | M1 | M2 | M3 | PwoC1 | PwoC2 | PwoC3 | Modification success | Classification accuracy |
|---|---|---|---|---|---|---|---|---|
| *Brisk* | 0 | NA | 1 | NA | 1 | NA | 67% | 64% |
| *ChatGPT* | 1 | 2 | 1 | 2 | 1 | 2 | 50% | 50% |
| *Claude* | 1 | 2 | 1 | 0 | 1 | 1 | 67% | 50% |
| *Coteach* | 0 | 2 | 1 | 1 | 2 | 2 | 33% | 75% |
| *DeepSeek* | 0 | 1 | 1 | 0 | 1 | 1 | 67% | 83% |
| *Gemini* | 1 | 1 | 1 | 1 | 0 | 1 | 83% | 58% |
| *Grok* | 1 | 0 | 0 | 0 | 0 | 1 | 33% | 67% |
| *Khanmigo* | 0 | NA | 1 | NA | 1 | 1 | 75% | 50% |
| *Magic School* | 0 | 1 | 1 | 1 | 0 | 1 | 67% | 67% |
| *Perplexity* | 1 | 1 | 2 | 1 | 0 | 1 | 67% | 58% |
| *School.AI* | 1 | 1 | 1 | 1 | 1 | 2 | 83% | 58% |
| *Modification Success by Task* | 55% | 56% | 82% | 56% | 55% | 70% | | |

*Note*: 0 = undershot, 1 = successfully met criteria of *PwC*, 2= overshot to *Doing Mathematics*

# Appendix E: Example AI-Modified Tasks

**ChatGPT's successful modification of the PwoC2 task:**

**Original Task:** *Use cross products to solve the proportion 4/96 = x/6000. Show your work.*

**ChatGPT's modification:**

*A map uses a scale where 2 inches represents 50 miles.*

*a. Use a diagram (tape diagram, line model, or scale drawing) to represent the scale and to show how you would find the number of miles represented by 7.5 inches.*

*b. Solve the problem using cross-products and by finding a unit rate. Show both methods and the steps you take.*

*c. Explain in words why the cross-product method and the unit-rate method give the same result. In your explanation, refer to your diagram and describe the multiplicative relationship between the numbers (i.e., how multiplicative reasoning about "how many times bigger" underlies the procedure).*

*d. Generalize: Write an algebraic statement that shows when two ratios (a/b = c/d) are equivalent, and use algebra to demonstrate why the cross-product step (ad = bc) is valid. Then create a different real-world proportion (not about maps) and show how the same reasoning applies.*

*(Show all work, include your diagram, and include at least one sentence that connects the procedure to the underlying proportional concept.)*

**Deep Seek's undershot modification of the PwoC1 task**

**Original Task:** *Convert the fraction 3/8 to a decimal and a percent. Show your work.*

**Deep Seek's modification of PwoC2 that undershot PwC criteria**:

*A student is looking at the fractions 3/8 and 5/8. They claim that since 3/5 is equivalent to 0.6, then 3/8 must be equivalent to 0.8.*

*Use a visual model (e.g., a diagram of a square, a number line, or a circle) to determine if the student's claim is correct. Explain your reasoning based on the model.*

*Now, use the standard division algorithm to convert 3/8 to a decimal. How does the result 0.375 connect to the visual model you created? What does each digit in 0.375 represent in terms of the whole?*

*A classmate states that 0.375 is the same as 37.5%. Explain why this makes sense by connecting the decimal to the idea of "per hundred."*

**Coteach's overshot modification of the M2 task**

**Original Task:** *True or False, 4/100 < 4/96*

**Coteach's modification of M2 that overshot PwC criteria**:

*The School Garden Project*

*Two classes are working on school garden projects:*
*Class A: Has completed 4 out of 100 planned garden squares*
*Class B: Has completed 4 out of 96 planned garden squares*

*Part 1: Initial Comparison*
*1. Without calculating decimal values, predict which class has completed a greater fraction of their garden project. Explain your reasoning using what you know about fractions.*

*2. Create visual models (such as grids or bar diagrams) to represent both fractions. What do you notice when you compare them visually?*

*Part 2: Mathematical Analysis*

*3. Use at least two different mathematical methods to determine whether 4/100 < 4/96. For each method, explain: Why the method works mathematically, how it connects to the meaning of fractions, what it reveals about the relationship between numerators and denominators.*

*4. Convert both fractions to equivalent fractions with the same denominator. Explain how this method helps you understand why one fraction is larger than the other.*
*Part 3: Pattern Recognition and Generalization*

*5. Consider the general case: If you have two fractions with the same numerator (like a/b and a/c), how can you determine which is larger without doing any calculations? Test your rule with other examples.*

*6. Explain why your rule works by connecting it to the real-world garden context. What does it mean for the denominator to be larger when the numerator stays the same?*

*7. Create your own comparison problem similar to this one and explain how someone could solve it using the insights you've developed.*